# BMJ Health & Care Informatics

# Exploring celebrity influence on public attitude towards the COVID-19 pandemic: social media shared sentiment analysis

Brianna M White [1], Chad Melton,[2] Parya Zareie,[1] Robert L Davis,[1] Robert A Bednarczyk,[3] Arash Shaban-Nejad [1]

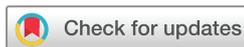

[1]College of Medicine, Department of Pediatrics, The University of Tennessee Health Science Center, Memphis, Tennessee, USA
[2]Bredesen Center for Interdisciplinary Research and Graduate Education, The University of Tennessee Knoxville, Knoxville, Tennessee, USA
[3]Hubert Department of Global Health, Rollins School of Public Health, Atlanta, Georgia, USA

**Correspondence to**
Brianna M White;
bwhite66@uthsc.edu

Professor Arash Shaban-Nejad;
ashabann@uthsc.edu

## ABSTRACT

**Objectives** The COVID-19 pandemic has introduced new opportunities for health communication, including an increase in the public's use of online outlets for health-related emotions. People have turned to social media networks to share sentiments related to the impacts of the COVID-19 pandemic. In this paper, we examine the role of social messaging shared by Persons in the Public Eye (ie, athletes, politicians, news personnel, etc) in determining overall public discourse direction.
**Methods** We harvested approximately 13 million tweets ranging from 1 January 2020 to 1 March 2022. The sentiment was calculated for each tweet using a fine-tuned DistilRoBERTa model, which was used to compare COVID-19 vaccine-related Twitter posts (tweets) that co-occurred with mentions of People in the Public Eye.
**Results** Our findings suggest the presence of consistent patterns of emotional content co-occurring with messaging shared by Persons in the Public Eye for the first 2 years of the COVID-19 pandemic influenced public opinion and largely stimulated online public discourse.
**Discussion** We demonstrate that as the pandemic progressed, public sentiment shared on social networks was shaped by risk perceptions, political ideologies and health-protective behaviours shared by Persons in the Public Eye, often in a negative light.
**Conclusion** We argue that further analysis of public response to various emotions shared by Persons in the Public Eye could provide insight into the role of social media shared sentiment in disease prevention, control and containment for COVID-19 and in response to future disease outbreaks.

### WHAT IS ALREADY KNOWN ON THIS TOPIC

⇒ Many studies have highlighted the persuasive nature of celebrity behaviour and messaging—both beneficial and detrimental to public health. As populations grow to trust the influential nature of celebrity activity on social platforms, followers are disarmed and open to persuasion when faced with the dissemination and rapid spread of misinformation. Exposure to large amounts of misinformation can have lasting effects on overall well-being and sociopsychological health.

### WHAT THIS STUDY ADDS

⇒ In our previous works, our team has successfully employed various natural language processing (NLP) models for the analysis of social media shared sentiment. With the utilisation of a fine-tuned DistilRoBERTa NLP model, sentiment and content analysis could uncover a correlation between COVID-19 related messaging shared by Persons in the Public Eye and public sentiment and discourse direction.

### HOW THIS STUDY MIGHT AFFECT RESEARCH, PRACTICE OR POLICY

⇒ Our discoveries could aid in better understanding public perception and attitude towards infectious disease mitigation efforts based on social influences, providing officials and policymakers tools to combat mis/disinformation shared via social media platforms and bolster disease prevention, control and containment for COVID-19 and in response to future disease outbreaks.

## INTRODUCTION

The COVID-19 pandemic has changed the lives of individuals and societies worldwide. Since its identification by the WHO in late 2019, the USA alone has since seen more than 87 million confirmed cases and 1 million virus-related deaths.[1] With millions of people forced out of public spaces, conversations surrounding the pandemic have taken place online, with a vast majority via social media networks.[2 3] People in the Public Eye (PIPE) such as news anchors, politicians, athletes and entertainers have taken this opportunity to discuss diverse topics with their platform followers, including their own experiences and opinion about health issues[3] such as COVID-19, COVID-19 vaccines and other public health measures. These shared experiences and opinions by PIPE often give way to public discourse, with many reacting to polarising statements by liking, commenting and resharing. As populations grow to trust the influential nature of celebrity activity on social platforms, followers are disarmed and open to persuasion when faced with false information, creating opportunities for dissemination and rapid spread of misinformation and disinformation.[4 5] Exposure to large amounts of misinformation can have lasting effects on overall well-being and sociopsychological health.[6] We argue this threat to population health should create a sense of urgency and warrants public health response to identify, develop and implement innovative mitigation strategies.

## Background

As social media use has sustainably grown amid the COVID-19 pandemic, public health researchers have capitalised on the potential for data mining of shared messages on social platforms.[7] Ease of access and rapid collection of data permits researchers to follow pandemic progression alongside online sentiment, providing tools for niche discovery and exploration of the emotion behind health decision making. For example, mining tweets from a specified period allows for parallel analysis of general public opinion during major events (ie, the release of new treatments such as vaccines or the death of a celebrity post-COVID-19 infection). Regarding COVID-19 vaccination specifically, researchers have used this recent increase in opinion sharing to measure overall sentiment and vaccine hesitancy or acceptance.[8–11] Moreover, many studies published over the past two decades highlight the persuasive nature of celebrity behaviour and messaging—both beneficial and detrimental to public health.[12–15] Others have used emotional diffusion networks to investigate the correlation between messaging shared by governmental agencies on social platforms and subsequent sentiment shared in response by the general public.[16] All present strong evidence in support of impacts on health-related perception, emotion and behaviour as a result of sentiment shared by those with societal influence or authority. Sentiment analysis is the practice of extrapolating the sentiment of a subject, idea, event or phenomenon by classification of written texts as some value of polarity (ie, positive or negative).[17] In our previous works,[18–20] our team has successfully employed various natural language processing (NLP) models for the analysis of social media shared sentiment. The application of such analytic tools could allow for time-expanded retrospective analysis of online sentiment in correlation with the progression of the ongoing pandemic. This targeted approach could provide tools for niche discovery and exploration of the emotions behind health decision making throughout the COVID-19 pandemic era and enhance preparedness, response and recovery efforts for future health crises.

Herein, we argue that, with the utilisation of a fine-tuned DistilRoBERTa NLP model,[18] sentiment and content analysis could uncover a correlation between COVID-19 vaccine-related messaging shared by PIPE and public sentiment and discourse direction. This discovery could aid in better understanding public perception and attitude towards vaccination based on social influences, providing officials and policymakers tools to combat mis/disinformation shared via social media platforms moving forward.

## METHODS
### Data

Our study compared COVID-19 vaccine-related Twitter posts (tweets) that co-occurred with mentions of PIPE ranging from 1 January 2020 to 1 March 2022. PIPE representatives were chosen due to publicly made statements that were either antivaccine in nature or identified as misinformation, as well as propagated misinformation. These individuals included Joe Rogan (JR), Tucker Carlson (TC), Nicki Minaj (NM), Aaron Rodgers (AR), Novak Djokovic (ND), Eric Clapton (EC), Rand Paul (RP), Phil Valentine (PV), Donald Trump (DT), Ted Cruz (TeC), Candace Owens (CO) and Ron DeSantis (RD). To evaluate and analyse differences in sentiment related to various figures, we divided the chosen PIPE representatives into three subgroups: (1) athletes/entertainers, (2) politicians, and (3) news personnel. We harvested approximately 13 million tweets with the Python library, Snscrape. For data cleaning, tweets by suspected bots, highly repetitive news media, highly repetitive-high frequency users or duplicates were removed. The data were then queried for the presence of COVID-19 vaccine-related terms and PIPE together. After the cleaning and querying process, our final data set consisted of 45 255 tweets composed of 34 407 unique authors. The Tweets contained a total of 16.32 million likes with a maximum of 70 228, an average of 45.76 an IQR of 2.0 and a median of zero.

### Sentiment classification

The sentiment was calculated for each tweet using a fine-tuned DistilRoBERTa model[10] that was created for a previous sentiment analysis study.[18] Pretrained models such as BERT, RoBERT and DistilRoBERTa are readily available as a free resource to researchers from platforms such as *huggingface.co*. This level of access and ability for rapid data collection determined our use of DistilRoBERTa, considering many other models take several days on dozens of tensor processing units (TPUs) to learn.[21 22] The model was trained by labelling approximately 4000 tweets as positive or negative. These labelled Tweets were then used with the method of back translation to create an augmented dataset of approximately 50 k tweets. The fine-tuning process achieved accuracy and an F1 score of approximately 0.96. After our model was fine-tuned, our data were processed with the Hugging Face pipeline for sentiment analysis. The model reported polarity (ie, positive or negative) as well as a probabilistic confidence score from [0,1]. Results will be discussed in terms of per cent positive (ie, n=positive tweets/total

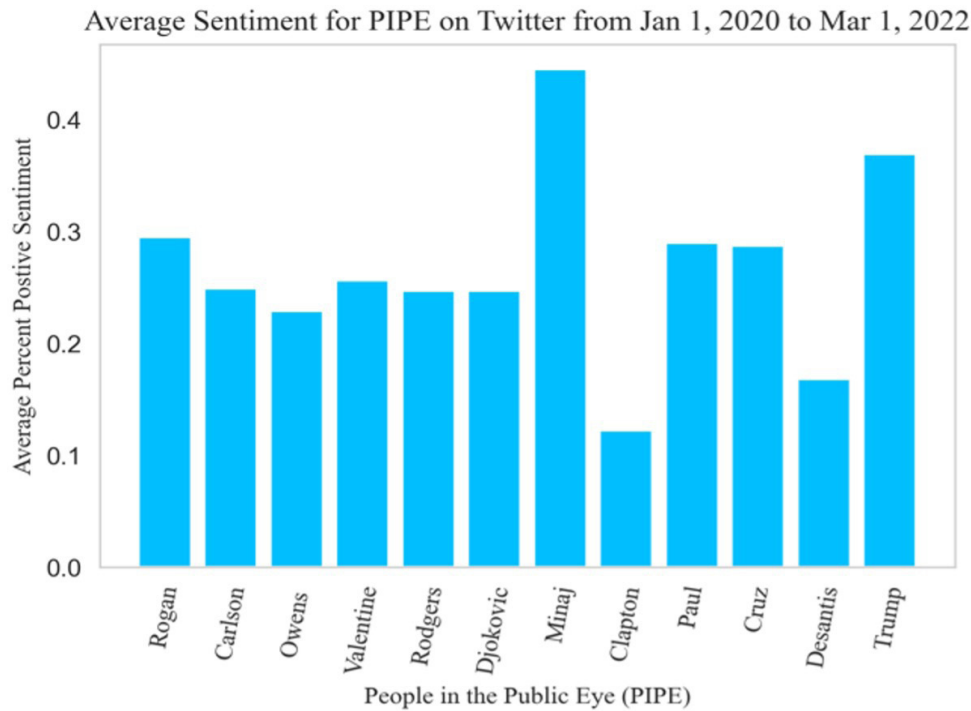

**Figure 1** An abstract representation of the average sentiment shared by People in the Public Eye (PIPE) via the social media platform, Twitter, from 1 January 2020 to 1 March 2022.

tweets) and per cent negative (ie, n=negative tweets/total tweets).

## RESULTS

As shown further, our study findings reflect public sentiment towards COVID-19 vaccines and vaccination in relation to messaging shared by chosen PIPE subgroups. Overall, though oscillatory over our specified period, the related sentiment was found to be overall more negative than positive (see figure 1).

### Athletes/entertainers

The maximum positive sentiment for tweets mentioning ND was 0.44 in February 2021, with a minimum of 0.1 in July 2021 and a mean of 0.25. There was a total of 29 210 likes on tweets mentioning ND, with a mean of 46.63. The average confidence score for all negative tweets mentioning ND was 0.96, while the positive tweets' score was 0.88. AR's sentiment maximum of 0.30 in December 2021, a minimum of 0.17 in February 2022 and a mean of 0.25. Tweets mentioning AR had a maximum of 29 210 likes with a mean of 59.56. Negative tweets mentioning AR had a mean confidence score of 0.95, while positive ones had a mean of 0.87. In June 2021, the EC's sentiment reached a max of 0.19, a min of 0.07 and a mean of 0.12. A total of 7327 people liked tweets EC was mentioned in and the mean was 23.82. Tweets mentioning EC labelled as negative had a mean score of 0.97, whereas the ones labelled as positive had a mean score of 0.84. NM sentiment maximum of 0.66 in May 2021, a minimum of 0.10 in February 2022 and a mean of 0.44. Tweets mentioning NM had a maximum of 16 165 likes and a mean of 39.19.

Negative tweets mentioning NM had an overall mean score of 0.93 and positive tweets' mean score was 0.87 (see figure 2).

### Politicians

When querying for social media posts related to politicians, results were noted to have large gaps in shared attitudes throughout our specified period. To begin, DT had a total of 11 195 tweets in our dataset. The maximum positive sentiment for DT occurred in February 2020, at 0.83 per cent positive, but saw a sharp decline a month later to 0.23 per cent positive, before remaining consistently low throughout the pandemic. The overall mean sentiment related to DT was determined to be 0.369, and tweets containing DT's name obtained a maximum of 52 650 likes, with a mean of 31.14. The dataset contained a total of 1520 tweets mentioning TeC. TeC-associated sentiment reached a maximum during May 2020 at 0.22, a minimum in January 2021 at 0.12 and a mean score of 0.287. Tweets mentioning TeC had a maximum of 70 228 likes and a mean of 74.51. References to RD occurred 3885 times in our dataset. Sentiments associated with RD reached a maximum during June 2021 at 0.966, a minimum of 0.25 in June 2020 and a mean of 0.168. Tweets mentioning RD had a maximum of 17 560 likes and a mean of 74.79. Lastly, Tweets mentioning RP totaled 2892 in our dataset, with a maximum sentiment of 0.6 in March 2020, a minimum of 0.16 in December 2021 and a mean of 0.290. Tweets mentioning RP had a maximum of 23 749 likes and a mean of 36.74 (see figure 3)

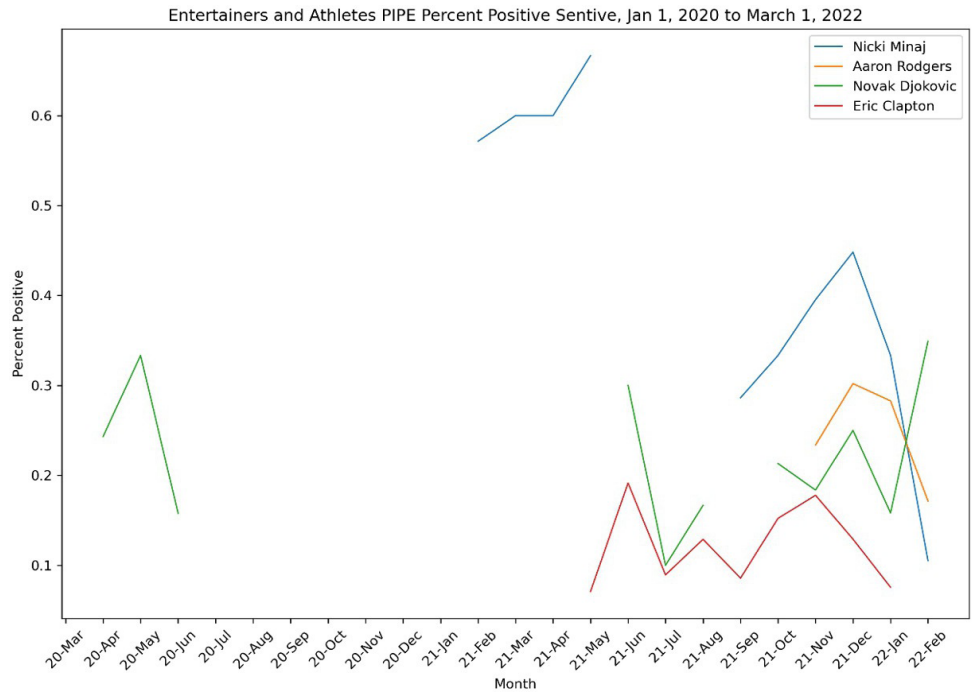

**Figure 2** Polarity per cent positive versus time. The blue line represents tweets mentioning Nicki Manaj. The orange line represents tweets mentioning Aaron Rodgers. The green line represents tweets mentioning Novak Djakovic. The red line represents tweets mentioning Eric Clapton.

### News personnel

For news media personalities (NMP), JR had a total of 6136 tweets in our dataset. The maximum positive sentiment for JR occurred in January 2021 but quickly reversed to reach a minimum in February 2021. The overall mean positive sentiment was determined to be 0.295. The mention of JR in a tweet received the maximum like count of 49 929 and a mean of 62.11. References to TC occurred 4843 times in our dataset. Positive sentiments associated with TC reached a maximum during April 2021 at 0.355, a minimum of 0.074 in August 2020 and a mean of 0.249. Tweets mentioning TC had a maximum of 24 586 likes and a mean of 39.35. PV had 1264 mentioning his name and reached his maximum sentiment of 0.4 in,[23] a minimum of 0.107 in December 2021 and a mean of 0.257. Tweets mentioning PV had a maximum of 12 264 likes and a mean of 62.75. Lastly, CO reached her maximum sentiment of 0.43 in January 2022, a minimum score of 0.038 in October 2021 and a mean of 0.229. Tweets mentioning CO displayed a maximum of 7195 likes and a mean of 38.60. CO occurred 996 times in our dataset. It is worth mentioning that occurrences of these personalities in tweets were not constant throughout the pandemic. TC and PV both were first mentioned in July 2020, JR in October 2020 and CO in November 2020 (see figure 4).

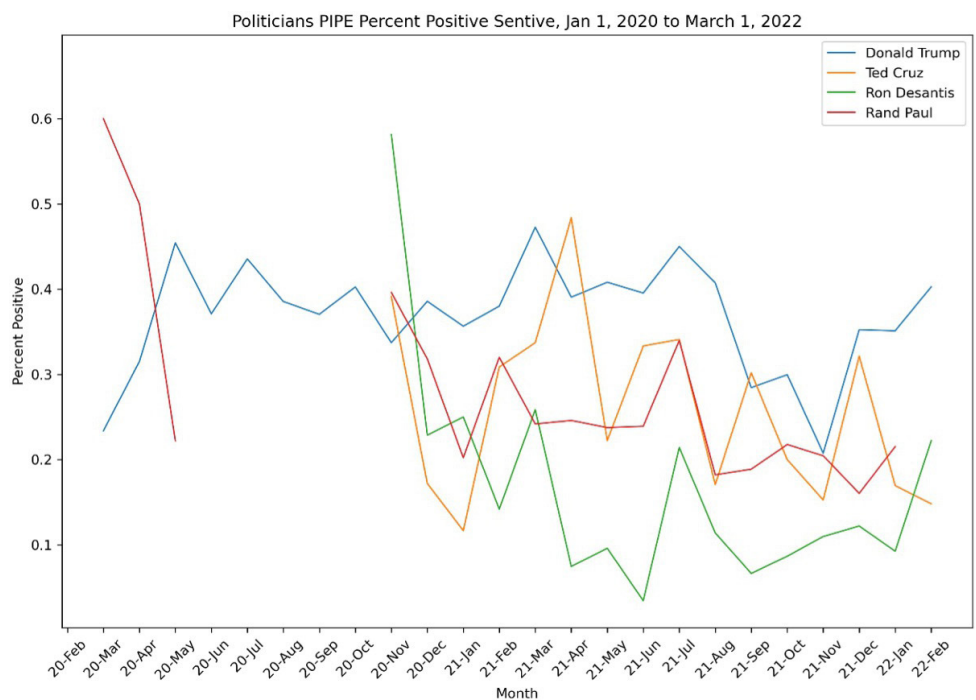

**Figure 3** Polarity per cent positive versus time. The blue line represents tweets mentioning Donald Trump. The orange line represents tweets mentioning TED cruz. The green line represents tweets mentioning Ron DeSantis. The red line represents tweets mentioning Rand Paul.

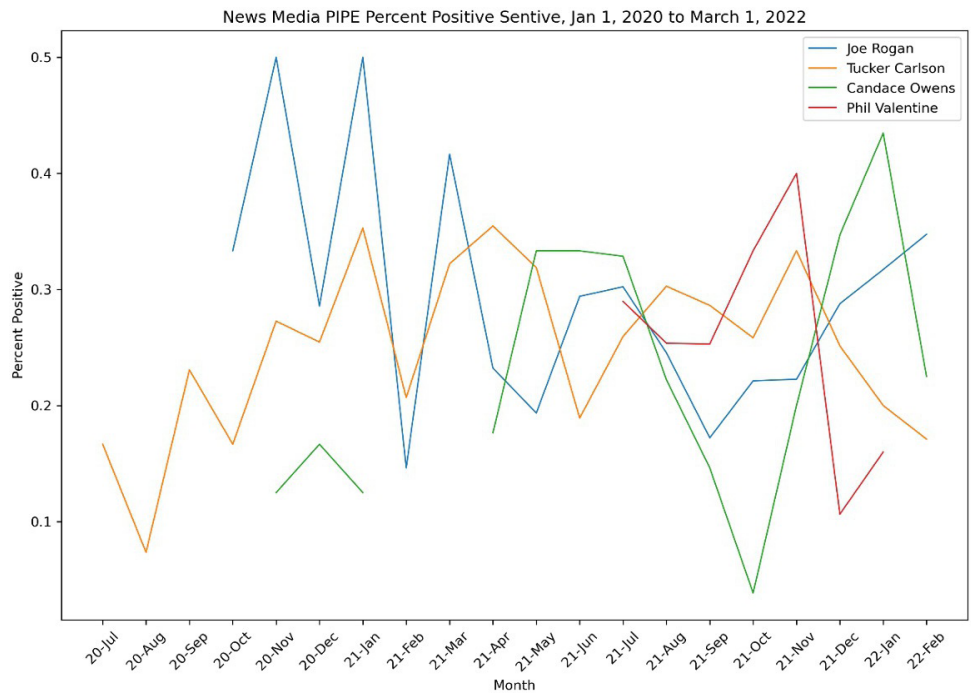

**Figure 4** Polarity percent positive versus time. The blue line represents tweets mentioning Joe Rogan. The orange line represents tweets mentioning Tucker Carlson. The green line represents tweets mentioning Candace Owens. The red line represents tweets mentioning Phil Valentine.

### Sentiment change analysis

Correlating changes in sentiment with real-world events can be challenging in large corpora without using computational topic modelling and/or semantic analysis. It is conceivable that some insight can be gained by noting the occurrence of sentimental shifts within a similar timeframe. With our data set, the majority of maximum or minimum sentiment occurred during different months. However, 3 months contained two occurrences of achieving a maximum or minimum sentiment (1 pos, 2 neg). NM and EC (both entertainers) achieved their maximum sentiment in June 2021. Following the approval of the Pfizer vaccine for ages 12–15 years, the only other major development was the detection of the delta variant. Two politicians (TeC and RP) expressed their most negative sentiment in December 2021. Many significant events occurred in December 2021 including the first detection of the COVID-19 omicron variant in the USA (1 December 2021), the Centers for Disease Control and Prevention (CDC) and FDA expanding booster recommendations to include everyone 16 years and older (9 December 2021), the reported death toll in the USA surpasses 800 000 (one in every 100 people ages 65+ years has died) (15 December 2021), the Pfizer-BioNTech or Moderna officially recommended over Johnson & Johnson vaccine (16 December 2021), the Omicron variant found to be 1.6 times more transmissible than Delta (20 December 2021) and the recommended isolation period shortened to 5 days on 27 December 2021. Two entertainers achieved their lowest sentiment in February 2022 (NM and AR). The only major events of February 2022 were related to a 30% rise in the death rate due to omicron infections. Furthermore, February 2022 was several months after NM and AR comments regarding vaccines. That being, it could be argued that sentiment related to entertainers or sports figures might not purely be driven by major developments in the pandemic.

### DISCUSSION

Our study examined the role of messaging shared by PIPE in determining general public discourse direction and the nature of sentiment shared via social media networks. Drawing on messaging shared by our three subgroups of influential users, findings suggest the presence of consistent patterns of emotional content shared by PIPE for the first 2 years of the COVID-19 pandemic influenced public opinion and largely stimulated online public discourse.

When comparing the overall mean score between negative and positive comments for individuals in the athletes/entertainers' subgroup, it can be concluded that related sentiment holds a more negative tone than positive. For this subgroup, the highest number of negative comments were associated with AR and NM, with a combined mean of 0.95. Regarding EC, it is interesting that he had very few positive tweets, the least in the group and the highest score difference. This fact implies that EC has been largely criticised by the public and was the least favourite figure of the subgroup. As evidence of this argument, the tweet with the highest like count mentioning EC states, '*Strongly disagree with [EC]…take on Covid and the vaccine and disgusted by his previous white supremacist comments. But if you reference the death of his son to criticize him, you are an ignorant scumbag*'.

The theme of overarching negative sentiment continues when examining findings within the politicians' subgroup. DT and TeC were found to have the most substantial impact within the subgroup, with a large number of combined likes totaling more than 122 000, with a monthly-based mean of 0.33. A substantial portion of tweets related to this subgroup was aimed at questioning whether politicians hold sufficient expert public health knowledge to advise constituents on medical decision making. This is reflected in the most liked tweet mentioning TeC, a user sharing, '*I called Ted Cruz's office asking to make an appointment to talk with the Senator about my blood pressure. They told me that the Senator was not qualified to give medical advice and that I should call my doctor. So I asked them to stop advising about vaccines*'. It became increasingly clear that this subgroup was the most tumultuous. The spread, reaction and engagement by the public to posts made by politicians online was indicative of a strong level of influence, suggesting politicians play key roles in ensuring population health and should be committed to promoting health-protective behaviours rather than sensational falsehoods.

As mentioned reviously, though sentiment related to the subgroup of news media personalities was shown to be oscillatory throughout the pandemic, emotion shared was overall more negative than positive. Moreover, tweets referencing this group were typically more related to antivaccine controversy or death (ie, death of PV) rather than news about vaccine development. Though the sentiment for the subgroup was overwhelmingly negative on average, a more interesting story begins to appear with the inspection of tweet content. For example, the most liked tweet associated with JR was '*I love how the same people who don't want us to listen to Joe Rogan, Aaron Rodgers about the covid vaccine, want us to listen to Big Bird & Elmo*', clearly a vaccine-hesitant or antivaccine statement. Notably, it is interesting to compare the number of tweets with the total number of maximum likes. This combined set of news media personalities had a total of 14 017 with 93 974 associated highest like count. The high number of likes displayed within these tweets shows that a much higher number of users are involved in reading tweets and are therefore potentially influenced by the content.

### Public health implications

We argue the application of our findings could have meaningful impacts on the public health sector, bolstering currently available surveillance tools for precision health promotion,[24] management of the ongoing COVID-19 pandemic and preparing for the next crisis. As we have demonstrated, messaging shared by influential members of society can have considerable effects on the directionality of public emotion and shared health decision making. Both negative and positive online social endorsement of prevention strategies such as vaccination are key in determining population-wide compliance and uptake success. However, threats of the spread of misinformation and disinformation by those with influence stand to undermine programmes supporting protective measures such as vaccination. As misinformation poses a range of psychological and psychosocial risks (anxiety, fear, etc), public health institutions hold some responsibility for the continued development and sustainability of low human effort surveillance systems optimised for generating responses to waves of falsities shared via social media platforms. Discernment of the dynamic levels of population sentiment shared via social media would allow public health officials to design catered mitigation and communication strategies. Social campaigns aimed at directing users with COVID-19 related inquiries to high-quality sources such as the CDC and other trusted public health institutions for evidence-based recommendations and instruction. Furthermore, public health and research institutions could be more proactive in creating collaborations with PIPE to share more positive messaging regarding vaccination. Moreover, there is room for intelligent algorithmic systems that identify patterns and anomalies in shared messaging, tasked with boosting messaging from social influencers who stand in affirmation of COVID-19 mitigation strategies; however, there is a need for a broader understanding of the potential negative implications, including ethical and legal issues.[23]

### Limitations

It is important to note some study limitations should be considered in unison with our findings. Sentiment analysis of social media shared messaging has long been challenging due to the ambiguity of natural language. Though language models such as BERT help to mitigate many of these challenges, the difficulty lies in the true detection of sarcasm, humour and complex inferences. As such, models currently available are unable to distinguish sentiment expressed towards different targets. As such, sentiment cannot be individually mined for two separate topics if shared within a short tweet. For example, if a user were to positively mention the name of a politician in support of COVID-19 vaccination while simultaneously sharing negative emotions toward communities in opposition, current language models may label the overall sentiment as negative. Though the negative emotion was not targeted to the mentioned politician, the model was limited in differentiating sentiment within the messaging. Moreover, though messaging shared by suspected bots, highly repetitive news media, highly repetitive high frequency users or duplicates were removed, it is possible a small amount could have still slipped through the data cleaning process.

## CONCLUSION

While faint differences in sentiment were observed across the subgroups, a broadly polarised negative tone was established. We demonstrate that as the pandemic progressed, public sentiment shared on social networks was shaped by risk perceptions, political ideologies and health-protective behaviours shared by PIPE. The risk of severe negative health outcomes increases with failure to comply with health-protective behaviour recommendations set forth by public health officials, such as vaccination, and our findings suggest that polarised messages from societal elites may downplay these risks, unduly contributing to an increase in the spread of COVID-19. In particular, we believe messaging shared by politicians and news personnel could be most strongly correlated with public health events; however, further experiments are needed. In the future, we plan to further explore the correlation of PIPE-shared sentiment with monumental events during the COVID-19 pandemic, such as the release of vaccinations or mask mandate liftings. Establishing a more thorough correlation between social media shared sentiment and health outcomes could help revolutionise public health responses to future infectious disease outbreaks.

**Contributors** BMW conceptualised and supervised the study and drafted, reviewed and edited the manuscript. CM and PZ conceptualised the study and drafted, reviewed and edited the manuscript. RLD reviewed and edited the manuscript. RAB reviewed and edited the manuscript. AS-N drafted, reviewed and edited the manuscript, supervised the study acting as guarantor, and acquired funding.

**Funding** This work was partially supported by National Cancer Institute (NCI) grant number #1R37CA234119-01A1.

**Competing interests** None declared.

**Patient consent for publication** Not applicable.

**Ethics approval** Not applicable.

**Provenance and peer review** Not commissioned; externally peer reviewed.

**Data availability statement** Data are available on reasonable request.

**ORCID iDs**
Brianna M White http://orcid.org/0000-0001-7576-5874
Arash Shaban-Nejad http://orcid.org/0000-0003-2047-4759